\documentclass[12pt]{article}

\usepackage[ruled,vlined]{algorithm2e}
\usepackage[english]{babel}
\usepackage[colorlinks=true, allcolors=blue]{hyperref}
\usepackage{bm}
\usepackage{tikz}
\usetikzlibrary{positioning}
\usetikzlibrary{shapes.geometric, arrows}
\usepackage{amsmath}
\usepackage{graphicx}
\usepackage{subcaption}
\usepackage{authblk} % Manages authors' affiliation
\usepackage{enumerate}
\usepackage{natbib} %comment out if you do not have the package
\usepackage{url} % not crucial - just used below for the URL 
\newcommand{\blind}{0}
\date{}

\DeclareMathOperator*{\argmax}{\text{argmax}}

\begin{document}

\def\spacingset#1{\renewcommand{\baselinestretch}%
{#1}\small\normalsize} \spacingset{1}

%%%%%%%%%%%%%%%%%%%%%%%%%%%%%%%%%%%%%%%%%%%%%%%%%%%%%%%%%%%%%%%%%%%%%%%%%%%%%%

\if0\blind
{
  
\title{\bf Stream-Based Active Learning for Process Monitoring}

\author[1]{Christian Capezza\thanks{Corresponding author. e-mail: \texttt{mailto:christian.capezza@unina.it}}}
\author[1]{Antonio Lepore}
\author[2]{Kamran Paynabar}

\affil[1]{Department of Industrial Engineering, University of Naples Federico II, Naples, Italy}
\affil[2]{School of Industrial and Systems Engineering, Georgia Institute of Technology, Atlanta, GA}

\setcounter{Maxaffil}{0}
\renewcommand\Affilfont{\itshape\small}
\date{}
\maketitle

} \fi

\if1\blind
{
  \bigskip
  \bigskip
  \bigskip
  \begin{center}
    {\LARGE\bf Stream-Based Active Learning for Process Monitoring}
\end{center}
  \medskip
} \fi

\bigskip

\begin{abstract}
Statistical process monitoring (SPM) methods are essential tools in quality management to check the stability of industrial processes, i.e., to dynamically classify the process state as in control (IC), under normal operating conditions, or out of control (OC), otherwise.
Traditional SPM methods are based on unsupervised approaches, which are popular because in most industrial applications the true OC states of the process are not explicitly known. 
% , and its knowledge is subject to non-negligible costs. 
This hampered the development of supervised methods that could instead take advantage of process data containing labels on the true process state, although they still need improvement in dealing with class imbalance, as OC states are rare in high-quality processes, and the dynamic recognition of unseen classes, e.g., the number of possible OC states.
This article presents a novel stream-based active learning strategy for SPM that enhances partially hidden Markov models to deal with data streams. 
The ultimate goal is to optimize labeling resources constrained by a limited budget and dynamically update the possible OC states. 
The proposed method performance in classifying the true state of the process is assessed through a simulation and a case study on the SPM of a resistance spot welding process in the automotive industry, which motivated this research.
\end{abstract}

\noindent%
\vfill

\newpage
\spacingset{1.45}

\section{Introduction}
\label{sec_intro}

In many modern industrial settings, data streams are generated continuously and used for statistical process monitoring (SPM) purposes. 
These data streams provide crucial insights to assess process stability, i.e., to identify whether a process is in an in-control (IC) or out-of-control (OC) state. 
In recent years, a growing body of the SPM literature has explored scenarios where labeled data are used to train supervised methods to distinguish between IC and OC states. 
\cite{zhang2015general} combined historical IC and OC data to set up a support vector machine model.
\cite{sun2023continual} proposed a general framework for classifying defects in batches of high volume data and continuously learning new types of defects, so leveraging the labeled data to improve the detection accuracy.
However, the time and costs associated with the labeling process can be extremely high and obtaining labels for all data points is often unfeasible.
Then, it is critical to develop strategies that can support decision on which data points should be annotated during a data stream by maximizing the labeling benefits while operating under a constrained budget. 
This requirement is relevant in manufacturing contexts, where labeling often relies on labor-intensive quality control inspections or costly equipment.

As an example, this need arises in the resistance spot welding (RSW) process in the modern automotive industry that motivated this research. 
In this context, it is critical to monitor the quality of welded spots during the body-in-white stage.
However, this requires expensive and time-consuming ultrasonic testing to confirm weld strength and detect defects like insufficient nugget formation or electrode wear \citep{xing2018quality} and makes it unfeasible to label all data.
On the other hand, dynamic resistance curve (DRC) measurements \citep{zhang2011resistance}, which are known to be proxy of the spot weld quality, are continuously collected at practically no cost in the form of data streams, thus motivating the need for an approach to select the most informative data points for labeling. 

Another example is in the semiconductor manufacturing process, where thousands of wafers are processed, and key metrics such as temperature, pressure, and chemical composition are continuously monitored. 
Determining the quality of each wafer often requires destructive testing, which is unfeasible at a large scale, further underscoring the need for intelligent labeling strategies \citep{shin2024framework}. 
Other examples are in chemical manufacturing or pharmaceutical quality control, which also generate data streams with prohibitive labeling costs and call for efficient labeling strategies in SPM scenarios.

To respond to this need, we propose to exploit active learning to offer an SPM framework for intelligently selecting the most informative data points to label in a stream-based scenario where labeling resources are scarce or expensive \citep{settles2012active}.
However, most active learning methods appeared in the literature focus instead on \textit{pool-based} sampling, which consists of selecting the most informative data points to label from a large static pool of unlabeled data, whereas in \textit{stream-based} active learning \citep{cohn1994improving,cesa2006worst} labeling decisions are made at the time of data acquisition. 
See \cite{cacciarelli2024active} for an extensive review of stream-based active learning.
More specifically, in this work, we focus on the single-pass setting, in which stream-based active learning algorithms evaluate each incoming data point on the fly to decide whether acquire its label.
The single-pass approach is particularly well suited for online Phase II SPM applications,  due to the immediate decision-making requirements, amd differs from batch-based variants, where the learner evaluates a fixed-size sample of data at a time and selects the most informative observations to be labeled \citep{zhang2020reinforcement}.

Although active learning has shown promise in various domains, its application to SPM remains relatively unexplored. 
Most existing stream-based active learning methods focus on independently and identically distributed data and only exploit the selected labeled instance information. 
That is, unlabeled data is often discarded, overlooking the opportunity given by the temporal dependencies inherent in data streams.
Some of the works that appeared in the literature (e.g., \cite{vzliobaite2013active,ienco2014high,mohamad2020online,liu2023online,schmidt2023streambased}) address the concept drift problem in stream-based active learning to capture the temporal evolution in the data distribution. 
However, these works are based on the assumption of class balance, which is unrealistic in high-quality processes where instead most of the data come from the IC process state, while data under OC conditions are very rare.
On the other hand, the stream-based active learning body of research that addresses the imbalance issue focuses only on binary classification \citep{chu2011unbiased} or does not consider concept drift \citep{carcillo2018streaming,zhang2018online}.
\cite{loy2012stream} consider a stream-based active learning setting with imbalanced classes and evolving data; however, they do not provide a budget management strategy that allows labeling the desired number of data points at the end of a data stream.

In the unsupervised context, it is worth mentioning hidden Markov models (HMMs) that have been used in the SPM literature for modeling data streams with unlabeled states, characterized by a multimodal distribution and temporal correlations \citep{wang2015hidden,alshraideh2014process}.
If only some observations can be labeled, HMMs can be generalized into partially hidden Markov models (pHMMs) to incorporate the knowledge of some observed process states  \citep{scheffer2001active,alemdar2017active}. 

The objective of this paper is to address the limited labeling resource constraint issue in SPM applications to monitor data streams characterized by temporal correlation, focusing on optimizing the selection of data points for annotation to maximize monitoring performance while minimizing labeling costs.
To achieve this, we propose a novel method for SPM that integrates a classifier based on pHMMs and a stream-based active learning strategy. 
The ability of pHMMs to capture temporal dynamics in data streams is combined with the efficiency of active learning to select informative data points.
The proposed method extends the active learning strategies proposed by \cite{scheffer2001active,alemdar2017active}, who have used pHMMs only in pool-based sampling settings.

The key contributions of this article can be summarized as follows: (i) we develop a novel stream-based active learning strategy tailored for process monitoring, which outperforms existing methods by effectively addressing challenges unique to process monitoring applications, such as concept drift and class imbalance; (ii) we employ a dual criterion for labeling to balance exploration and exploitation, where the exploitation criterion aims to label samples where the pHMM indicates high uncertainty, i.e., near the decision boundary, while the exploration criterion seeks to label samples that might reveal previously unobserved states; (iii) we develop an algorithm able to rapidly determine whether to label a new sample and update the estimated pHMM.

The rest of the article is organized as follows.
Section \ref{sec_overview} presents the basic principles and notation for HMMs and pHMMs and builds the proposed stream-based active learning strategy.
In Section \ref{sec_sim}, an extensive simulation study compares the proposed method with alternative approaches available in both the SPM and the active learning literature.
In Section \ref{sec_real}, the practical applicability of the proposed method is illustrated through a case study in SPM of a resistance spot welding (RSW) process in the automotive industry.
Section \ref{sec_conclusions} concludes the article. 
All computations and plots were obtained using the programming language \textsf{R} \citep{r2024}.

\section{Methodology}
\label{sec_overview}

\begin{figure}[t]
    \centering
    \includegraphics[width=\textwidth]{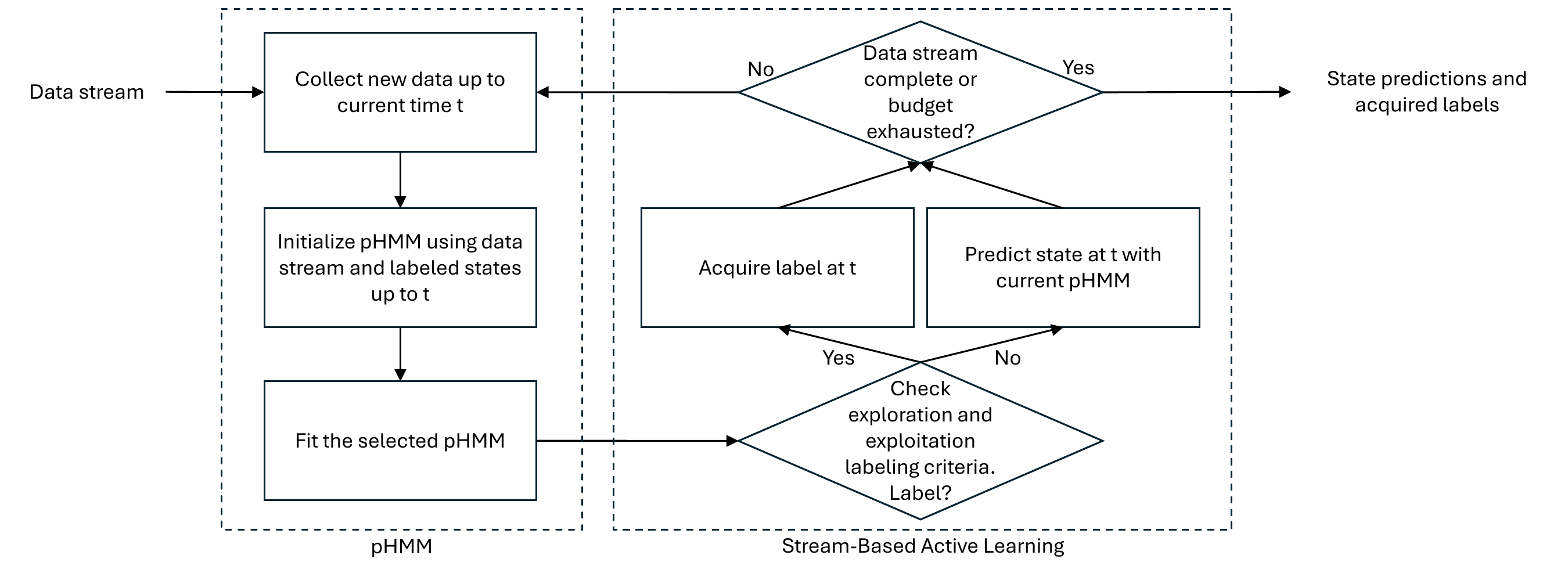}
    \caption{\label{fig_flowchart}
    Flow diagram of the proposed method.}   
\end{figure}
A flow diagram of the proposed method is shown in Figure \ref{fig_flowchart}.
In short, a pHMM is first initialized using the collected data stream up to time \( t \), along with previously labeled states. 
Then, the selected pHMM is fitted to these data. 
The methodology employs a stream-based active learning approach, where the exploration and exploitation criteria are evaluated to decide whether labeling is necessary at time $t$. 
If the labeling criteria are met, the label, i.e., the current state of the process, is acquired.
Otherwise, it is predicted according to the currently fitted pHMM.
This loop continues until the data stream is complete or the budget is exhausted. 
Basic principles and notation on HMMs and pHMMs are given in Section \ref{sec_hmm}, while Section \ref{sec_sbal} presents the stream-based active learning strategy.

\subsection{Partially Hidden Markov Model}
\label{sec_hmm}

Before introducing a pHMM, let us briefly introduce the basic concepts and notation of HMMs. Readers may refer to \cite{rabiner1989tutorial} for a comprehensive tutorial on HMMs.
Let us denote by $\bm y_t = (y_{t1}, \dots, y_{tp})^\top$ the measurement of a multivariate $p$-dimensional quality characteristic acquired at time $t$ and by $x_t$ the corresponding process state, with $t=1,2\dots,T$, where $T$ denotes the length of the sequence. 
The set of target states is denoted by $S=\lbrace 1, \dots, N \rbrace$, with $x_t = 1$ indicating that the process is IC at time $t$, or alternatively, $x_t = j \in \lbrace 2, \dots, N \rbrace$ to indicate that the process has transitioned to the corresponding OC state. 
Multiple states could also be considered for the IC process, as done by \cite{grasso2017phase} in the context of monitoring multimode processes.
Consequently, $\bm{Y}_t = ( \bm y_1, \bm y_2, \ldots, \bm y_t )^\top$ denotes the $t \times p$ matrix containing the sequence of the observations of the multivariate quality characteristic observations up to time $t$, while $\bm{x}_t = (x_1, x_2, \ldots, x_t)^\top$ denotes the corresponding sequence of process states up to time $t$.

In an HMM, the system being modeled is assumed to follow a Markov process, i.e., 
\begin{equation*}
p(x_{t+1} = i | \bm x_{t}) = p(x_{t+1} = i | x_{t}), \quad i=1,\dots,N, \quad t=1,2,\dots, 
\end{equation*}
where the process state $x_t$ is a latent variable, i.e., it is not observable directly. 
However, for inference purposes, an observable process $\bm y_t$ is assumed to be available that depends only on $x_t$, i.e., 
$$p(\bm y_t | \bm x_{t}, \bm y_{t-1}) = p(\bm y_t | x_t).$$
\begin{figure}[t]
    \centering
    \begin{center}
\begin{tikzpicture}[
  node distance=1.5cm,
  state/.style={circle, draw, minimum size=1.5em},
  observation/.style={rectangle, draw, minimum size=1.5em}
]
% States
\node[state] (s1) {$x_1$};
\node[state, right=of s1] (s2) {$x_2$};
\node[state, right=of s2] (s3) {$x_3$};
\node[state, right=of s3] (s4) {$x_4$};
% Observations
\node[observation, below=of s1] (o1) {$\bm y_1$};
\node[observation, below=of s2] (o2) {$\bm y_2$};
\node[observation, below=of s3] (o3) {$\bm y_3$};
\node[observation, below=of s4] (o4) {$\bm y_4$};
% Transitions
\draw[->] (s1) -- (s2);
\draw[->] (s2) -- (s3);
\draw[->] (s3) -- (s4);
% State to Observation
\draw[->] (s1) -- (o1);
\draw[->] (s2) -- (o2);
\draw[->] (s3) -- (o3);
\draw[->] (s4) -- (o4);
\end{tikzpicture}
\caption{Graphical representation of a hidden Markov model.\label{fig_HMM}}
\end{center}
\end{figure}
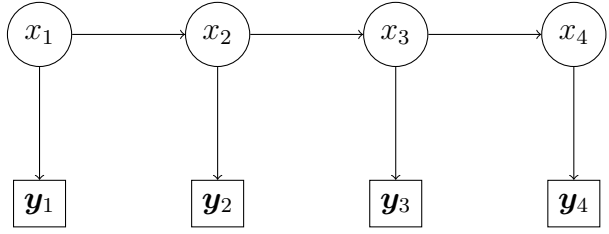
This is graphically represented in Figure \ref{fig_HMM} for $t=1,2,3,4$.
The parameters of an HMM are $\bm \theta = (\bm \pi, \bm A, \bm b)$, where $\bm \pi = (\pi_1, \dots, \pi_N)^\top$ is the initial state distribution, i.e., $\pi_i = p(x_1 = i)$, $i = 1,\dots, N$, $\bm A$ is the $N \times N$ transition probability matrix, with elements 
\begin{equation}
\label{eq_transition_prob}
a_{ij} = p(x_t = j | x_{t-1} = i), \quad \sum_{j=1}^N a_{ij} = 1, \quad i,j=1,\dots,N,
\end{equation}
and $\bm b = (b_1, \dots b_N)^\top$ are the emission distributions, i.e., $b_i(\bm y_t) = p(\bm y_t | x_t = i)$.
In this paper, we assume $b_i(\bm y_t) = \phi(\bm y_t; \bm \mu_i, \bm \Sigma_i)$, where $\phi(\cdot; \bm \mu, \bm \Sigma)$ is the probability density function of the $p$-dimensional Gaussian distribution with mean $\bm \mu$ and covariance matrix $\bm \Sigma$.

A pHMM is an HMM in which some of the true values of the state $x_t$ (labels) are revealed.
In particular, this work assumes that any $x_t$ label can be theoretically observed, but this information comes at a given cost, which is set equal for all labels without loss of generality.
The total cost cannot exceed a limited budget, expressed in terms of the proportion $0 < B < 1$ of the total number of samples.
Deferring the choice of which are the optimal $x_t$ labels to observe to Section \ref{sec_sbal}, the problem here is to build an $N$-class classifier able to predict the current state $x_t$ given the historical data acquired up to time $t$, consisting of fully observed $\bm Y_t$ and partially observed states $\bm x_t$. 

This problem can be reduced to the estimation of a pHMM, as was similarly done by \cite{scheffer2001active}, who consider a semi-supervised learning setting where $\bm x_t$ is known for a subset of samples. \cite{scheffer2001active} developed a constrained version of the Baum-Welch algorithm, which is a special case of the expectation-maximization algorithm used to estimate the unknown parameters of an HMM.
This algorithm was recently improved by \cite{li2021new}.
The constrained Baum-Welch algorithm for estimating a pHMM is reviewed in Section \ref{sec_fit_hmm}, and Section \ref{sec_model_selection} discusses how to select the number of states $N$, i.e., model selection.

\subsubsection{Constrained Baum-Welch algorithm}

\label{sec_fit_hmm}

To consider partial labels, we define a vector of actions $\bm l = (l_1, \dots, l_T)^\top$, where $l_t = 1$ if the $t$-th sample is labeled and $l_t = 0$ otherwise.

\paragraph{Initialization}
Initialize the transition probabilities $\bm A$ and initial state probabilities $\bm \pi$.
Initialize also the parameters of the emission distributions $b_i$ for $i=1,\dots,N$, i.e., the means $\bm \mu_i$ and the covariances $\bm \Sigma_i$.
See Appendix \ref{sec_init} for more details.

\paragraph{Expectation step (E-step)}
Compute the forward probabilities $\alpha_t(i) = p(x_t = i, \bm y_t; \bm \theta)$ using the \textit{forward algorithm}:

\begin{enumerate}
\item For $t=1$, initialize the forward probabilities using the initial state distribution and the emission probabilities as $\alpha_1(i) = \pi_i b_i(\bm y_1)$, for $i = 1, \ldots, N$.

\item For $t = 2, \dots, T$ and $i=1,\dots,N$, compute the forward probabilities recursively:
\begin{equation}
\label{eq_alpha}
\alpha_t(i) = 
\begin{cases}
\phi(\bm y_t; {\bm \mu}_i, {\bm \Sigma}_i) \sum_{k=1}^{N} \alpha_{t-1}(k) a_{ki} & \text{if } l_t = 0 \text{ or } (l_t = 1 \text{ and } x_t = i), \\
0 & \text{if } l_t = 1 \text{ and } x_t \neq i.
\end{cases}
\end{equation}

\end{enumerate}
Note that the likelihood is obtained as $l(\bm \theta) = p(\bm Y_T; \bm \theta) = \sum_{i=1}^N \alpha_T(i)$.
Then, compute the backward probabilities $\beta_t(j)=p( \bm y_{t+1}, \dots, \bm y_T| x_t = i; \bm \theta)$ using the \textit{backward algorithm}:

\begin{enumerate}
\item Initialize the backward probabilities $\beta_T(i)=1$ for $i=1, \ldots, N$.

\item For $t = T-1$ to $1$, compute the backward probabilities recursively, for $i=1,\dots,N$:
\begin{equation}
\label{eq_beta}
\beta_t(i) = 
\begin{cases}
\sum_{j=1}^{N} \beta_{t+1}(j) a_{ij} \phi(\bm{y}_{t+1}; \bm{\mu}_j, 
\bm{\Sigma}_j) & \text{if } l_t = 0 \text{ or } (l_t = 1 \text{ and } x_t = i), \\
0 & \text{if } l_t = 1 \text{ and } x_t \neq i.
\end{cases}
\end{equation}
\end{enumerate}
Based on $\alpha_i(t)$ and $\beta_i(t)$ calculated in Equation \eqref{eq_alpha} and Equation \eqref{eq_beta}, compute the probability $\gamma_i(t)$ of being in state $i$ at time $t$ given the entire observation sequence $\bm Y_T$:

\begin{equation}
\label{eq_gamma}
    \gamma_t(i) = p(x_t = i | \bm Y_T; \bm \theta) = \frac{p(x_t = i, \bm Y_T; \bm \theta)}{p(\bm Y_T; \bm \theta) } = \frac{\alpha_i(t) \beta_i(t)}{\sum_{i=1}^N \alpha_i(t) \beta_i(t)}, \quad i=1,\dots,N.
\end{equation}
The next step is to compute for $i=1,\dots,N$ the probability $\xi_t(i,j)$ of being in state $i$ at time $t$ and state $j$ at time $t+1$, given the entire observation sequence $\bm Y_T$:

\begin{equation}
\label{eq_xi}
\begin{aligned}
\xi_t(i,j) = p(x_t = i, x_{t+1} = j | \bm Y_T; \bm \theta) 
= \frac{p(x_t = i, x_{t+1} = j , \bm Y_T; \bm \theta)}{p(\bm Y_T;\bm \theta)} =  \\
=\frac{\alpha_i(t) a_{ij} \beta_j (t+1) \phi(\bm y_{t+1}; \bm \mu_j, \bm \Sigma_j)}{\sum_{k=1}^N \sum_{h=1}^N \alpha_h(t) a_{kh} \beta_k (t+1) \phi(\bm y_{t+1}; \bm \mu_k, \bm \Sigma_k)},  \quad i, j = 1, \ldots, N.
\end{aligned}
\end{equation}

\paragraph{Maximization Step (M-step)}
Using Equation \eqref{eq_gamma} and \eqref{eq_xi}, update the initial state probabilities for each $i=1,\dots,N$ as $\pi_i = \gamma_1(i)$, then update the transition probabilities in Equation \eqref{eq_transition_prob} as 
\begin{equation*}
a_{ij} = \frac{\sum_{t=1}^{T-1} \xi_t(i,j)}{\sum_{t=1}^{T-1} \gamma_t(i)}, \quad i,j=1,\dots,N.
\end{equation*}
Update the Gaussian emission distribution parameters as 
\begin{equation*}
\bm \mu_i = \frac{\sum_{t=1}^T \gamma_{t} (i) \bm y_t}{ \sum_{t=1}^T \gamma_t(i)}, \quad \bm \Sigma_i = \frac{\sum_{t=1}^T \gamma_{t} (i) (\bm y_t - \bm \mu_i) (\bm y_t - \bm \mu_i)^\top}{ \sum_{t=1}^T \gamma_t(i)}, \quad i=1,\dots,N.
\end{equation*}

\textbf{Iteration}.
Repeat the E-step and M-step until the parameter estimates converge, or until a certain number of iterations is reached.

The constrained Baum-Welch algorithm produces the estimate of the pHMM parameters $\hat{\bm \theta} = (\hat{\bm \pi}, \hat{\bm A}, \hat{\bm b})$ and can then be used as a probabilistic classifier for a new observation through conditional probability $p(x_{T+1} = i | \bm Y_{T+1}; \hat{\bm \theta})$, which can be easily calculated with the forward-backward algorithm as the quantity $\gamma_{T+1}(i)$ using Equation \eqref{eq_gamma}.

A fundamental challenge in fitting a pHMM is initialization.
The Baum-Welch algorithm is guaranteed to converge only to a local maximum.
In this article, we propose a new algorithm for the initialization of the pHMM that exploits the prior information available on the problem.
Since the majority of the samples come from an IC process, a robust estimate of location and scatter provides the estimate of the IC process state parameters, while we initialize the OC state parameters by looking at the parts of the sequence that show the largest deviation from the IC state distribution.
The algorithm is presented in detail in Appendix \ref{sec_init}.

\subsubsection{Model selection}
\label{sec_model_selection}

The model selection in the context of pHMMs pertains to determining the number $N$ of all possible states/classes. 
The number of observed states in the training sample only determines a lower bound for $N$.
Information-based criteria allow for the selection of a model based on the principle of parsimony, which results from the trade-off between model fit and model complexity.
Several works have explored the performance of information criteria for HMMs \citep{celeux2008selecting,costa2010model}.
We propose employing the Akaike information criterion (AIC, \cite{akaike1974new}) for model selection, defined as $\text{AIC} = - 2 l({\hat{\bm \theta}}) + 2 k$, where $l({\hat{\bm \theta}})$ denotes the maximized log-likelihood and $k$ the number of free parameters.
%is $k = N-1 + N(N-1) + Np + Np(p+1)/2$.
In this article, we consider a parsimonious parameterization, suitable for the highly imbalanced number of the states, which assumes that all states share a common covariance matrix $\bm \Sigma_1 = \dots = \bm \Sigma_N = \bm \Sigma$, then 
\begin{equation}
\label{eq_aic}
\text{AIC} = - 2 l({\hat{\bm \theta}}) + 2 \left[ N-1 + N(N-1) + Np + p(p+1)/2 \right].
\end{equation}
At each time $t$, given the available data, we fit the pHMMs at different numbers of possible states. 
Subsequently, the model with the minimum AIC is selected as the current classifier. 
If some states become observable after labeling, the minimum number of possible states is updated to reflect the number of observed states.

\subsection{Stream-Based Active Learning}
\label{sec_sbal}

Stream-based active learning aims to provide a strategy for making decisions at time $t$ based on current measurements $\bm Y_{t}$ and partially observed states $\bm x_t$.
These decisions include determining the state of the process as IC, i.e., $\hat{x}_{t}=1$, identifying an OC state $\hat{x}_{t}=j \in \lbrace 2, \dots, N \rbrace$, or requesting the label $x_{t}$ if the sample has high uncertainity.
This work proposes a novel stream-based active learning strategy for process monitoring that incorporates temporal dynamics in the labeling strategy and employs a dual criterion strategy to effectively balance exploration and exploitation. Considering parts of the sequence rather than the current samples also improves the labeling strategy.
The exploitation criterion aims to label samples close to the decision boundary, where the pHMM indicates high uncertainty. In contrast, the exploration criterion seeks to label samples that might reveal previously unobserved states.
These criteria are detailed in Sections \ref{sec_exploitation} and \ref{sec_exploration}.
The budget is divided between the two labeling strategies by two positive weights $w^\text{exp}$ and $w^{\text{exr}}$ that sum to one.
More sophisticated choices could be based on contextual bandit approaches \citep[e.g.,][]{wassermann2019ral,zeng2023ensemble}.
However, the proposed approach has shown satisfactory results and has the advantage of being simple and computationally fast, which is crucial in a stream-based setting for real-time implementation.
The methodology of the proposed active learning method is described in a succinct way in Algorithm \ref{algo:active_learning}.
\begin{algorithm}
\caption{Stream-Based Active Learning with pHMM}\label{algo:active_learning}
\spacingset{1}
\SetKwInOut{Input}{Input}
\SetKwInOut{Output}{Output}
\SetArgSty{textnormal} % Set the style of algorithmic arguments

\Input{Initial data stream of IC data 
$D^{\text{init}}$, data stream \( D \), budget \( B \) \\
Weight of exploration $w_{\text{exr}}$ and exploitation $w_{\text{exp}} = 1 - w_{\text{exr}}$\\
Weighting parameter of the MEWMA statistic $\lambda$}
\Output{State predictions $\hat{\bm x}_t = (\hat{x}_1, \dots, \hat{x}_T)^\top$, set of selected labels $L$} 

Fit the pHMM with one state on $D^{\text{init}}$, initialize \( L = \emptyset \), $b = \lfloor B \cdot T \rfloor$; \\

\For{\( t \) from $T_{\text{init}} + 1$ to \( T_{\text{init}} + T \)}{
    Set $B_t =  b / (T + T_{\text{init}} - t + 1) $, $B^{\text{exr}} = w_{\text{exr}} B_t$, $B^{\text{exp}} = w_{\text{exp}} B_t$\;
    Fit pHMM with data stream up to $\bm y_t$\; 
    Produce a point prediction $\hat x_t$ and a probabilistic prediction of $x_t$\;
    Calculate the entropy \eqref{eq_entropy} associated to the prediction and its $p$-value $p^{\text{exp}}$\;
    \If{
    $p^{\text{exp}} < B^{\text{exp}}$ and $B_t < B$}
    { $\hat x_t = x_t$, $L = L \cup \lbrace t \rbrace$, $b = b - 1$
    }
    \Else{
      Calculate the MEWMA monitoring statistic \eqref{eq_ewma_monitoring} and its $p$-value $p^{\text{exr}}$;\\
      \If{$p^{\text{exr}} < B^{\text{exr}}$ and $B_t < B$ and $\hat{x}_t > 1$}
      { $\hat x_t = x_t$, $L = L \cup \lbrace t \rbrace$, $b = b - 1$
    }
    }
}
\Return{$\hat{\bm x}_t$, $L$;}
\end{algorithm}

The proposed algorithm requires, as input, an initial data stream assumed to be acquired from an IC process $D^{\text{init}} = \lbrace (
x_{1} = 1, \bm y_1), \dots, (x_{T^{\text{init}}} = 1, \bm y_{T^{\text{init}}}) \rbrace$.
The availability of a small dataset $D^{\text{init}}$ is reasonable and serves to establish an initial estimate of the IC state distribution parameters \citep{trittenbach2021overview}. 
Then, together with the data stream $D=\lbrace \bm y_{T_{\text{init}} + 1}, \dots, \bm y_{T_{\text{init}} + T} \rbrace$ subject to the labeling decision, the other inputs of the algorithm are the budget $B \in [0,1]$, i.e., the proportion of samples to be labeled, the weights $w_{\text{exr}} \in [0,1]$ and $w_{\text{exp}} = 1 - w_{\text{exr}}$, and the weighting parameter $\lambda$ for the multivariate exponentially weighted moving average (MEWMA) statistic described in Section \ref{sec_exploration}.

The algorithm is initialized with the pHMM fitted on $D^{\text{init}}$, starting with an empty set of labeled samples $L = \emptyset$, and sets the total number of available labels $b = \lfloor B \cdot T \rfloor$.
As the observation $\bm y_t$ arrives, the budget is updated to $B_t=b/(T+T_{\text{init}}-t+1)$ and then divided between the exploration and exploitation criteria, according to their respective weights, as $B^{\text{exr}} = w_{\text{exr}} B_t$ and $B^{\text{exp}} = w_{\text{exp}} B_t$.
Then, the estimate of the pHMM parameter $\hat{\bm \theta}$ is updated with the new observation $\bm y_t$.
Note that the number of states $N$ is selected based on the AIC in Equation \eqref{eq_aic}.
The prediction of the state is performed using the updated pHMM, providing a probabilistic prediction using Equation \eqref{eq_gamma} as $\gamma_t(i) = p(x_t = i | \bm Y_t; \hat \theta)$, for each state $i=1,\dots,N$, and a point prediction is determined as $\hat x_t = \argmax_{i=1,\dots,N} \gamma_t(i)$.

The algorithm decides whether to trust the model prediction $\hat x_t$ or to request a label to refine the pHMM and improve future classifications. 
This decision is made according to the exploration and exploitation criteria aforementioned.
If neither criterion justifies labeling, the sample remains unlabeled, and $\hat x_t$ is used for state classification.
Otherwise, the label is acquired, $\hat x_t$ is updated to the observed value, and the count of available labels $b$ is reduced by one.
Since the labeling decision is made when the sample is acquired, a critical aspect of stream-based active learning is ensuring that the label spending remains within the predetermined budget.
As in \cite{liu2023online}, we add the condition that labeling is allowed only if the estimated budget does not exceed the available budget $B_t < B$. 
This ensures that our labeling strategy remains sustainable throughout data-stream processing.
The algorithm continues until all samples are processed ($t=T_{\text{init}} + T$), providing in the final output the state predictions and the set of labeled samples.

\subsubsection{Exploitation Criterion}
\label{sec_exploitation}

At each time step \( t \), after fitting a pHMM, we obtain a probabilistic prediction of \( x_t \) as \( \gamma_t(i) = p(x_t = i \mid \bm{Y}_t; \hat{\bm \theta}) \) for each state \( i = 1, \ldots, N \). 
The exploitation criterion focuses on labeling samples where the classifier exhibits the greatest uncertainty. 
This uncertainty is quantified using the entropy, defined as 
\begin{equation}
\label{eq_entropy}
H(x_{t} \mid y_{t}) = -\sum_{i=1}^N \gamma_{t}(i) \log(\gamma_{t}(i)).
\end{equation}
A threshold is then established such that, if the entropy exceeds this limit, a label is deemed necessary. 
This threshold is crucial because it must comply with the constraints on budget spending. 
We employ a parametric bootstrap approach to determine an appropriate threshold, simulating numerous sequences from the fitted pHMM. 
The procedure that simulates a sequence from a fitted pHMM is detailed in Appendix \ref{sec_sim_fitted}.
For each simulated sequence, the corresponding entropy is calculated. 
These calculated values are then used to set the \( p \)-value, denoted as \( p^{\text{exp}} \), which represents the proportion of simulated entropy values that exceed the observed entropy. 
If the uncertainty of the model about the current observation is excessively high,  \( p^{\text{exp}} \) will be small.
If \( p^{\text{exp}} < B^{\text{exp}} \), the exploitation criterion asks for the label.
Finally, it should be noted that, if the fitted pHMM has only $N=1$ state, there is no uncertainty about the classification and the exploitation criterion concludes that no label is needed.

\subsubsection{Exploration Criterion}
\label{sec_exploration}

Exploration is crucial in balancing the sampling bias inherent in querying labels predominantly for observations near decision boundaries. 
The exploitation criterion alone may fail to identify new, unobserved states if their emission distributions are far from regions of high uncertainty. 
The active learning literature offers various exploratory criteria, such as \cite{vzliobaite2013active}, which show that a simple random query strategy can be effective when combined with uncertainty-based criteria. 
However, the efficacy of random exploration has been demonstrated primarily in balanced classification settings. 
Discovering new OC states can be particularly challenging in SPM applications with a pronounced imbalance between the IC and OC states. 
Thus, it is essential to allocate the labeling budget efficiently, prioritizing, in the exploration criterion, samples that may represent new classes.
Additionally, leveraging the temporal dynamics within the sequence allows us to borrow strength from consecutive observations, facilitating the identification of new OC states.

We begin with the pHMM model fitted to the available data up to the current time \( t \), with the selected number of states $N$. 
For every state $i=1,\dots,N$, we compute the MEWMA statistic for the current observation \( \bm y_{t} \) as:
\begin{equation*}
\bm z_{ti} = \lambda (\bm y_{t} - \hat{\bm \mu}_i) + (1 - \lambda) \bm z_{(t-1)i},    
\end{equation*}
If the true state at time $t$ is $x_t=i$, the steady-state distribution of the MEWMA statistic can be estimated as \( N(\bm{0}, \lambda / (2 - \lambda) \hat{\bm \Sigma}_i) \) and we can calculate the squared statistical distance of $\bm z_{ti}$ from its mean as
\begin{equation}
\label{eq_ewma_i}
V^2_{ti} = \frac{2 - \lambda}{\lambda} \bm z_{ti}^\top \hat{\bm \Sigma}_i^{-1} \bm z_{ti}.
\end{equation}
If the observation $\bm y_t$ comes instead from a new unseen state, all these distances should be large.
Then, we define the following monitoring statistic 
\begin{equation}
\label{eq_ewma_monitoring}
V^2_t = \min_{i=1,\dots,N} \frac{2 - \lambda}{\lambda} \bm z_{ti}^\top \hat{\bm \Sigma}_i^{-1} \bm z_{ti}.
\end{equation}
If the current observation is part of a sequence originating from a new unseen OC distribution, \( V_t^2 \) is expected to be significantly large. 
Consequently, the exploration criterion asks for labeling if \( V^2_t \) is too large according to budget constraints. 
We calculate the \( p \)-value, denoted as \( p^{\text{exr}} \), as the probability of observing values of the chi-square distribution with \( p \) degrees of freedom $\chi^2_p$ that exceed $V^2_t$. 
If \( p^{\text{exr}} < B^{\text{exr}} \), the exploration criterion asks for the label.
Although there are more sophisticated methods in the EWMA literature for estimating the distribution of statistics \( V^2_t \), this approach has numerically proven to be effective.

\section{Simulation Study}
\label{sec_sim}

We present an extensive simulation study to assess the performance of the proposed method in monitoring data streams.
In Section \ref{sec_sim_comparsion} the proposed method is compared with alternative approaches available in the literature, while Section \ref{sec_sim_weight} deals with the choice of the weight of exploration and exploitation.

\subsection{Comparison with Alternative Approaches}
\label{sec_sim_comparsion}

To simplify the comparison with alternative approaches without loss of generality, the process is assumed to have three possible states, i.e., $S = \lbrace 1, 2, 3 \rbrace$. 
In particular, $x_t=1$ if the process is IC, $x_t=2$ if it is in the first OC state, and $x_t = 3$ if it is in the second OC state.
The measurement of the multivariate quality characteristic is a $p$-dimensional random vector, where we consider three values of $p \in \lbrace 10, 20, 30 \rbrace$.
When the process is IC, the emission distribution $p(\bm y_t | x_t = 1)$ is set equal to $b_1 (\bm y_t) = \phi(\bm 0, \bm \Sigma)$, where the $(i, j)$-th element of the covariance matrix $\bm \Sigma$ is $\sigma_{ij} = 0.75^{|i - j|}$, $i,j=1,\dots,p$.
Then, we set $p(\bm y_t | x_t = 2) = b_2 (\bm y_t) = \phi(\bm d_1, \bm \Sigma)$ and $p(\bm y_t | x_t = 3) = b_3 (\bm y_t) = \phi(\bm d_2, \bm \Sigma)$, where we set the $p$-dimensional vectors $\bm d_1 = (\delta, 0, \dots, 0)^\top$ and $\bm d_2 = (0, \delta, 0, \dots, 0)^\top$, with $\delta$ being a positive scalar denoting the size of the shift from the IC process mean, and is made varying in the set $\lbrace 0, 0.6, 1.2, 1.8, 2.4, 3 \rbrace$.
Since we assume that the covariance matrices are equal for all states, we are interested in identifying process mean shifts.

The data generation process is described below.
Given a simulated state sequence $x_1 , x_2, \dots$, the corresponding multivariate quality characteristic observations $\bm y_1, \bm y_2, \dots$ are simulated from the corresponding emission distributions.
We describe how to simulate the sequence of states as follows.
Each data stream is initialized with $T^\text{init}=100$ observations in which the IC state 1 is known.
Then, an additional sequence of $T=500$ states is simulated, alternating a sequence of IC (state 1) values, with a length randomly sampled every time from 60 to 85, and a sequence of five OC state values.
Although an HMM assumes that the duration of time spent in a particular state follows a geometric distribution, the sequence lengths generated in this way are more realistic in practical applications while not affecting the generality of the results.

\begin{figure}[t]
    \centering
\includegraphics[width=0.8\textwidth]{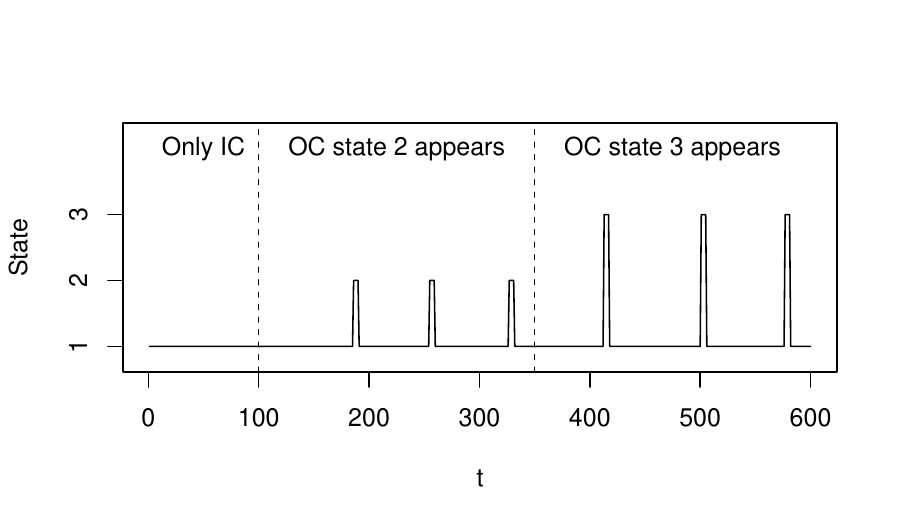}
    \caption{\label{fig_seq}
    Example of a simulated sequence of states.
    The state sequence is divided into three parts.
    The first part ($t \in \lbrace 1,\dots,T_{\text{init}}=100 \rbrace$) contains only initial IC state 1 values, in the second part ($t \in \lbrace 101,\dots, 350 \rbrace$) the process transitions to OC state 2, in the third part ($t \in \lbrace 351,\dots,600 \rbrace$) the process transitions to a new OC state 3.
    A line connects all the simulated states.}   
\end{figure}
During the first half (250 points) of the data stream, i.e., for $t=101,\dots,350$, OC values are generated from OC state 2, while for the second half, i.e., for $t=351, \dots, 600$, i.e., OC values are generated from OC state 3, as depicted in the example of a simulated sequence in Figure \ref{fig_seq}.
This generation pattern serves to assess the exploration ability of an SPM method in detecting OC states that have not been observed previously and to adapt to changing conditions effectively.

To compare the performance of the proposed method, we consider the following alternatives available in the literature.
As a classical SPM competitor, we examine a standard MEWMA control chart \citep{lowry1992multivariate}, hereinafter referred to as \textit{MEWMA}. 
This is an unsupervised method, since it does not use information from the OC state observations.
The MEWMA monitoring statistic $V^2_t$ has already been introduced in Section \ref{sec_exploration} to introduce the exploration labeling criterion.
Here, we use it for classification purposes.
Assuming that most observations come from an IC process with distribution \( N(\bm \mu_1, \bm \Sigma_1) \), robust estimates \( \hat{\bm \mu}_1 \) and \( \hat{\bm \Sigma}_1 \) can be used to minimize the influence of OC states. 
Then we can calculate the monitoring statistic $V^2_{t1}$ as in Equation \eqref{eq_ewma_i}.
When $V^2_{t1}$ is smaller than the upper control limit, the state of the process is classified as IC; otherwise, it is classified as OC.
Another unsupervised alternative is an HMM fitted to the data sequence with no access to the state labels and hereinafter referred to as \textit{unsupervised}. 
The HMM autonomously determines the number of states by selecting the model that minimizes the AIC across all numbers of states. Subsequently, it classifies the state at time $t$ based on the current model.
Among active learning strategies, we consider a baseline approach that randomly selects which labels to acquire, hereinafter referred to as \textit{random}, and a more sophisticated method of requesting labels, which we refer to as \textit{equispaced}, that asks for a label every $1 / B$ samples to cover the data stream evenly and, if there is a correlation among consecutive states, to exploit neighboring observations.
The \textit{proposed} method is implemented as in Section \ref{sec_sbal} and integrates both exploitation and exploration within the stream-based active learning strategy.

In the simulation study, the budget $B$ is set to assume five equally spaced values between 0.01 and 0.2.
To assess the effectiveness of these methods in accurately classifying the state of the process, we use the F1 score, which is calculated as the harmonic mean of the precision and recall metrics.
Precision represents the ratio between true positive (OC state) predictions and the total number of positive predictions, while recall is the ratio of true positive predictions to the total number of actual positive instances.
Thus, the F1 score measures the ability to identify both IC and OC conditions while considering the class imbalance.
For each shift size and budget value, we conducted 48 simulation runs and calculated the average F1 score achieved by each method. 
It is trivial to remark that when the labeling strategy asks for a label, the corresponding observation will be correctly classified. This implies that the more a sample is difficult to classify, the greater the reward.

\begin{figure}[t]
    \centering
    \includegraphics[width=.95\textwidth]{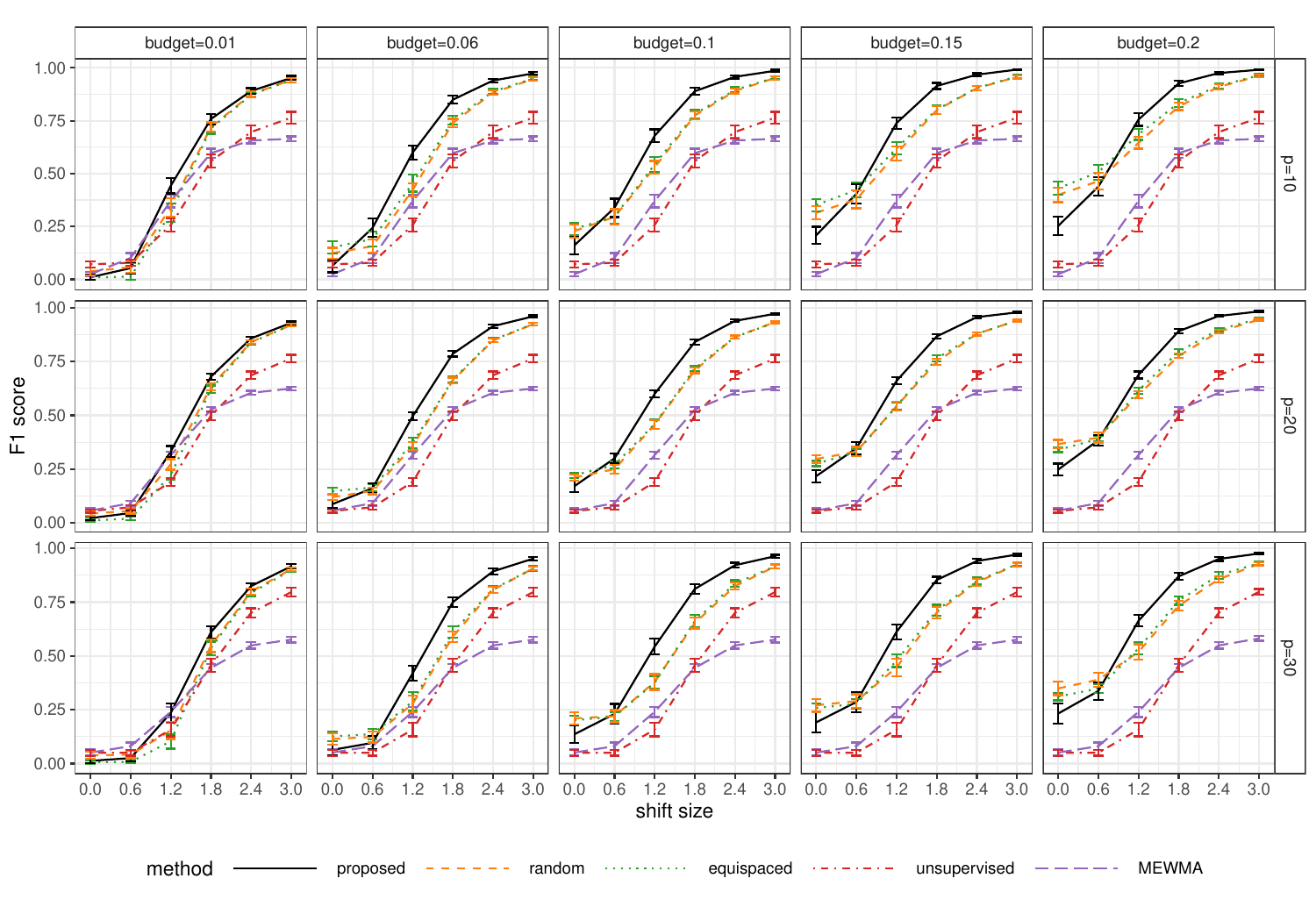}
    \caption{\label{fig_shift}
    F1 score plotted as a function of the shift size for each available budget and number of variables.}   
\end{figure}
In Figure \ref{fig_shift}, we report the average F1 score achieved as a function of the shift size for all the competing methods over the 48 simulation runs, with error bars allowing uncertainty quantification. 
%obtained by adding $\pm 2 \hat \sigma_{\text{F1}}$ to the average F1 score, where $\hat \sigma_{\text{F1}}$ is the estimated standard deviation of the mean F1 score.
Each row of panels corresponds to a different number of variables $p$, while each column corresponds to a different active learning budget $B$.
Since \textit{MEWMA} and \textit{unsupervised} do not depend on the labeled samples, their results do not vary across columns.
The performance of the \textit{MEWMA} method is markedly unsatisfactory. 
This outcome is anticipated since this approach solely examines the in-control (IC) distribution of the data. 
The \textit{unsupervised} approach is comparable to \textit{MEWMA}, with a slightly better performance at a larger shift size, which is expected. 
A large shift indeed allows the unsupervised HMM to identify OC states and then use their information more easily, even without using labels.

As expected, all active learning methods are at least as good as \textit{unsupervised} because they use labels. In addition, they exhibit a general improvement in classification performance as the budget increases at every mean shift level.
Regarding the baseline active learning strategies, \textit{random} and \textit{equispaced} exhibit similar performance.
Between the two, we expected a better performance of \textit{equispaced} at higher budgets if it is able to populate the data stream and identify all shifts adequately.
However, \textit{equispaced} shows a minor improvement over \textit{random}, only at the largest budget value. 

The \textit{proposed} method outperforms all competitors,
even though it provides a slight improvement at very low budget values ($B=0.01$). 
Its efficiency in utilizing labels for both exploitation and exploration leads to a progressively greater advantage over competitors as the budget grows.
At the largest budget values, the \textit{proposed} method maintains its edge over competitors, albeit slightly diminishing. 
This trend is expected since, with increasing budgets, all active learning strategies eventually encounter all OC states. 
The \textit{proposed} method excels then for intermediate budget values (for $B$ between 0.06 and 0.15) and process mean shift sizes (for $\delta$ between 1.2 and 2.4).

Figure \ref{fig_shift} allows us to also discuss the effect of the number of variables $p$ on the performance of the methods. 
We observe a slight decrease in performance for each method as $p$ increases, which is expected since the larger the number of variables, the more difficult the estimation of model parameters.
The main results discussed above are still valid for every value of $p$.

\subsection{Choice of the Exploration and Exploitation Weights}
\label{sec_sim_weight}

The proposed stream-based active learning method uses the weight parameters $w_{\text{exr}}$ and $w_\text{exp} = 1 - w_{\text{exr}}$, between zero and one, to balance exploration and exploitation. 
For example, the choice $w_{\text{exp}}=w_{\text{exr}}=0.5$ gives equal importance to the two criteria.
In this section, we evaluate the impact of varying these weights on the performance of the proposed method.
Specifically, we consider $w_{\text{exp}}$ to assume the following values: 0 (i.e., only exploration), 0.25, 0.5, 0.75, and 1 (i.e., only exploitation).
\begin{figure}[t]
    \centering
    \includegraphics[width=.95\textwidth]{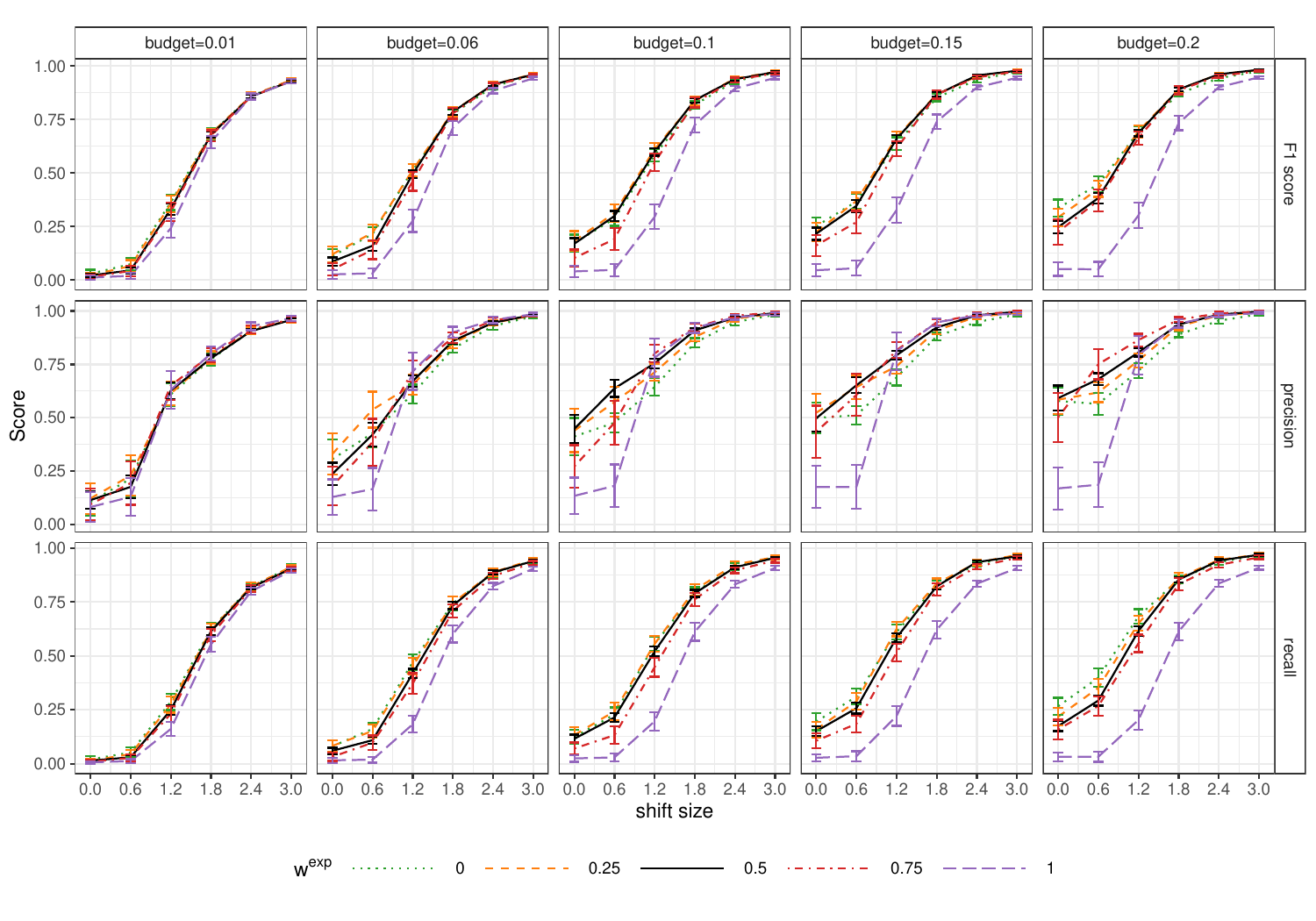}
    \caption{\label{fig_w}
    F1 score, precision, and recall scores achieved with the proposed method, plotted as a function of shift size, for each available budget and $p=20$. Each line corresponds to a given weight of the exploitation criterion $w^{\text{exp}}$.}   
\end{figure}
This analysis is conducted under the same conditions of Section \ref{sec_sim_comparsion}, but limited to $p=20$.
The results are illustrated in Figure \ref{fig_w}, which shows the classification performance across different $w_{\text{exp}}$ values, with each column of panels representing varying budget value and each row corresponding to a different classification metric.

The first row of panels shows the performance of the F1 score.
In particular, assigning full weight to exploitation results in the poorest performance across all budget levels. 
In contrast, full exploration yields a better F1 score by helping to detect new OC states. 
The intermediate settings, which integrate both exploration and exploitation, generally perform better, with similar results across these configurations. 

It is worth investigating the precision and recall scores separately for a deeper understanding.
These metrics are depicted in the second and third rows of Figure \ref{fig_w}, respectively. 
The recall score helps explain the poor results when relying solely on exploitation. 
Without exploration, after identifying the initial OC state, the method fails to recognize new unseen OC states emerging in the data stream, leading to many missed positives.
Although recall alone might suggest that complete exploration is sufficient, precision analysis generally prefers methods with a heavier exploitation component, except in cases of pure exploitation at smaller shift sizes. 
Exploitation focuses on labeling the most uncertain samples, refining decision boundaries, and hence increasing the likelihood that positive predictions are correct.

This analysis underscores the importance of using both exploration and exploitation. 
Although $w_{\text{exp}} = w_{\text{exr}} = 0.5$ is a solid starting point, further refinements could enhance the F1 score.

\section{Case Study}
\label{sec_real}

The case study described in this section concerns the monitoring of an RSW process in the automotive body-in-white manufacturing industry \citep{zhang2011resistance}. 
RSW is a prevalent method for joining, through two copper electrodes, overlapping galvanized steel sheets, and ensuring the structural integrity and solidity of the welded assemblies in vehicles \citep{martin2014assessment}. 
The quality characteristic monitored in this case study is the dynamic resistance curve (DRC), which is 
% a profile acquired practically without cost during the RSW process and is 
known to be related to the physical and metallurgical development of the corresponding spot weld and, thus, to its final quality of the joint produced \citep{capezza2021functional_clustering}.
Data for this study were collected by Centro Ricerche Fiat during lab tests on various car bodies of the same model. 
Each body featured a large number of spot welds, differing in metal sheet thickness, material, and welding duration. However, in this case study, we investigate DRC observations pertaining to a specific spot weld location across different car bodies. The DRC raw measurements, from which we get the smooth profiles, were collected at regular intervals of 1 millisecond.

\begin{figure}[t]
  \centering
  \begin{subfigure}[b]{0.4\textwidth}
    \includegraphics[width=\textwidth]{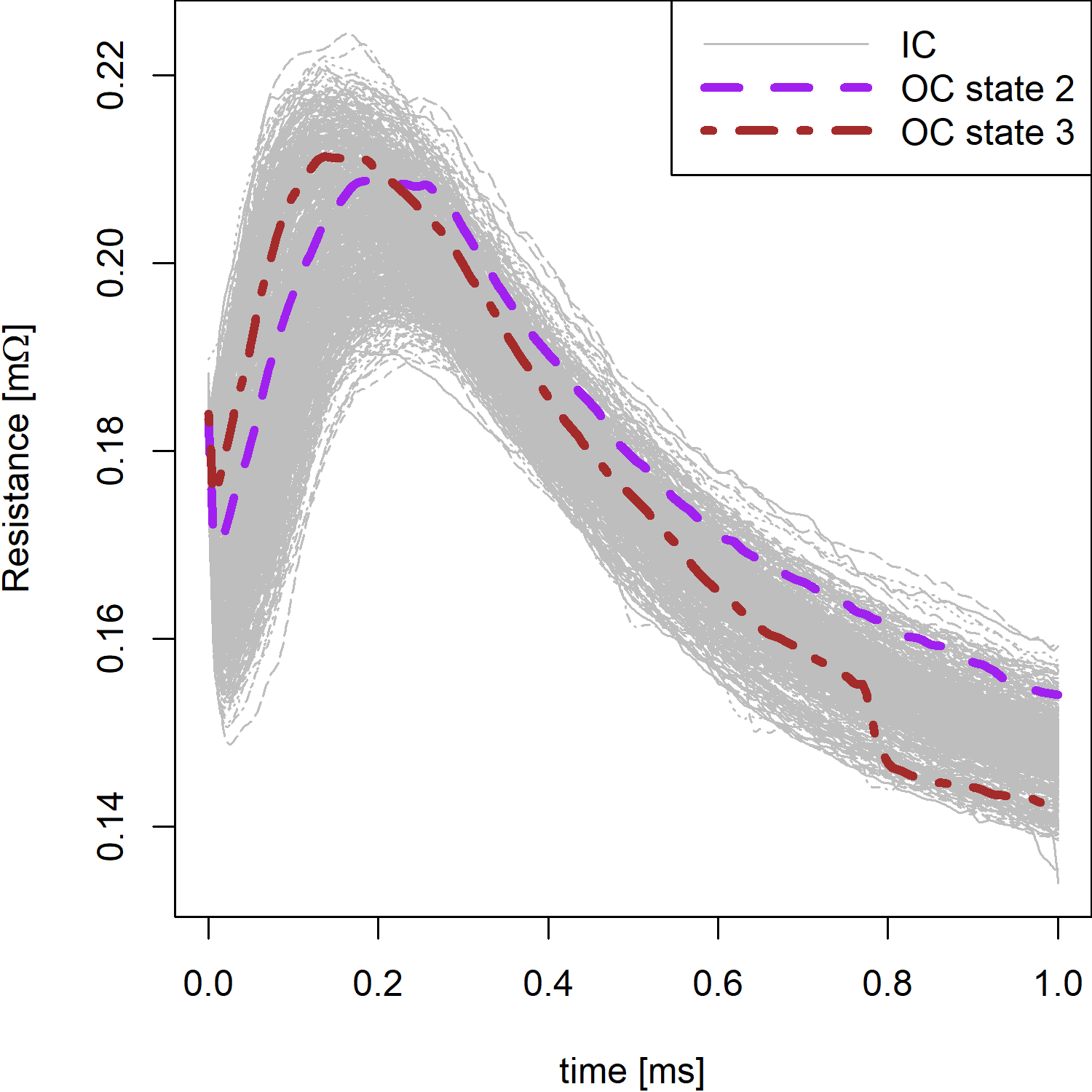}
    \caption{\label{fig_drc}}
  \end{subfigure}
  \hfill % this command puts space between the subfigures
  \begin{subfigure}[b]{0.5\textwidth}
    \includegraphics[width=\textwidth]{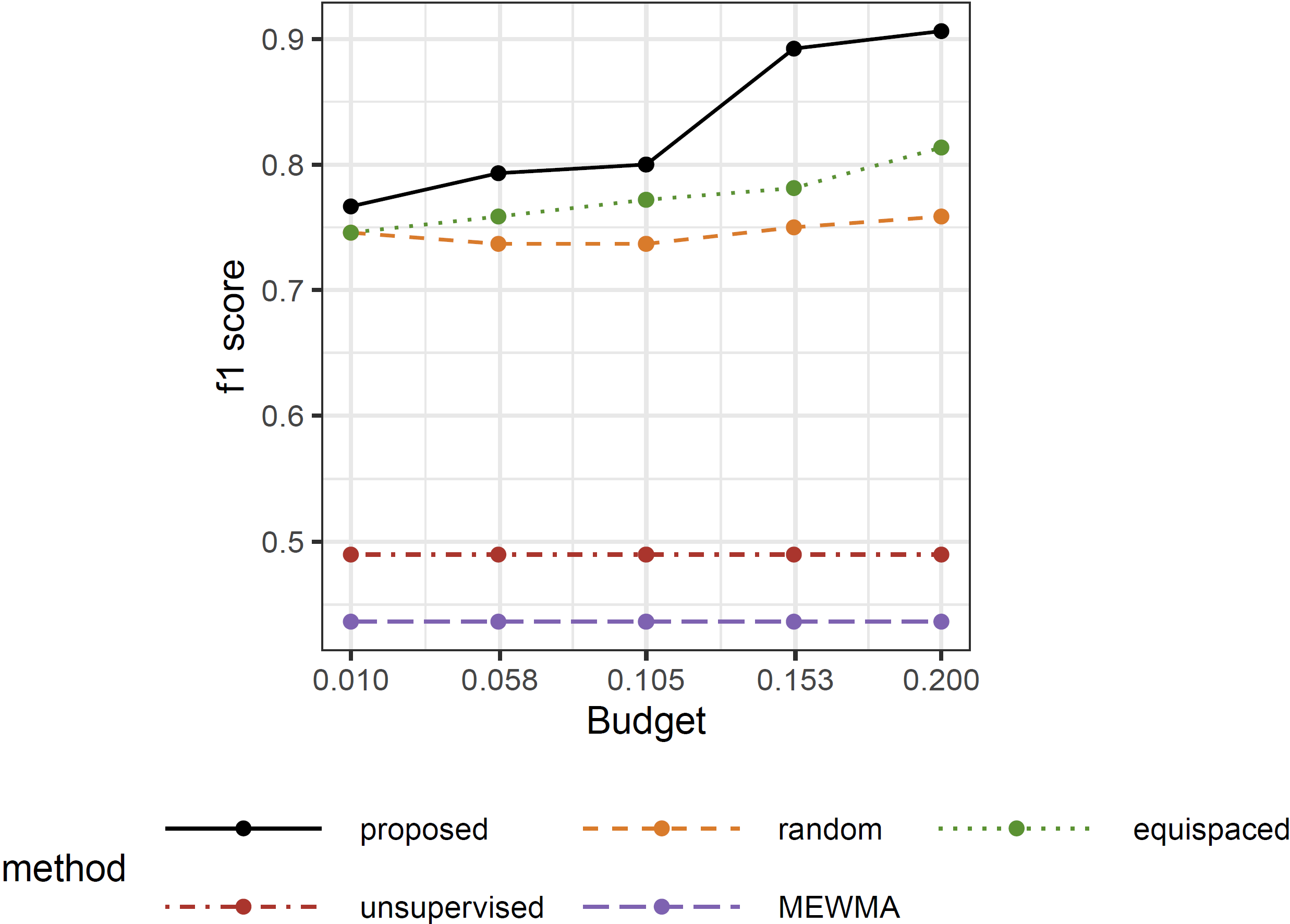}
    \caption{\label{fig_real}}
  \end{subfigure}
    \caption{(a) Plot of DRCs acquired during the resistance spot welding process on a specific spot weld produced for different car bodies.
    (b) F1 score plotted as a function of the shift size for each available budget. Each line corresponds to a given weight of the exploitation criterion.}
\end{figure}
Figure \ref{fig_drc} shows 600 DRCs corresponding to as many car bodies produced under the IC process state, hereinafter referred to as state 1.
In addition, the two DRCs superimposed on this figure, which display distinct anomalies, were known to pertain to two different OC states of the process, as certified by the ultrasonic test. 
In general, as also happens for this case study, spot welds from an RSW process are not routinely certified as IC or OC because the ultrasonic tests needed to do that are costly and time-consuming and cannot cover all samples.
In other words, while it is possible at a non-negligible cost, DRCs do not always have labels to indicate the true state ($x_t$) of the RSW process that has generated them.
Since DRCs can be instead routinely acquired in-line practically without costs, this situation strongly motivates the proposed stream-based active learning method, which could be extremely beneficial by directing costly and time-consuming ultrasonic tests only toward samples that can enhance the process monitoring performance.

As documented in the literature on RSW processes \citep{xing2018quality}, 
the dashed purple DRC displayed in Figure \ref{fig_drc} exhibits higher final resistance values, possibly indicating welds characterized by inadequately formed weld nuggets, sometimes referred to as cold welds.
We denote the corresponding state of the process as OC state 2.
The dash-dotted brown DRC instead is characterized by abrupt changes in resistance.
The sudden decrease in resistance during the welding process indicates the unwanted expulsion of molten material from the weld.
We denote the corresponding state of the process as OC state 3.

This case study aims to further compare the proposed method in a real scenario and demonstrate its practical applicability, although the advantages of the proposed method have already been highlighted in the simulation study presented in Section \ref{sec_sim}.
To mimic an RSW process lively streaming the DRC data presented above, as if the labels on the process true state were unknown but could be asked, at a certain cost, anytime for any profile, we generate a live sequence of 600 states $x_t$, where $x_t=1$ stands for IC, and $x_t=2$ and $x_t=3$ correspond to OC states 2 and 3, respectively. 
The first 100 $x_t$ are set equal to 1 (IC), while the remaining 500 are drawn from $\lbrace1,2,3\rbrace$, as in Section \ref{sec_sim}.
For each $x_t$, we sample (without replacement) a DRC observation from the 600 available. If $x_t=1$, the DRC observation is left as is, while if $x_t=j$, $j=2,3$, the sampled (IC) DRC observation is shifted by the difference $y_{OC_j}-m_1$, where $m_1$ is the average of the 600 IC DRC observations, and $y_{OC_j}$ represents the DRC observation displayed in Figure \ref{fig_drc} characterizing the OC state $j$.
The first 100 IC DRCs, due to the functional nature of the quality characteristic in this case study, are used for the typical pre-processing step \citep{ramsay2005functional} needed to obtain functional data from the raw profile measurements. 
For this purpose, a functional principal component analysis was performed and 10 functional principal components were retained to account for more than 99\% of the sample total variance.
These principal components remain unchanged in the subsequent stage, e.g., to project the remaining 500 DRC observations and extract scores for the pHMM estimation.
Discussion of different pre-processing techniques is beyond the scope of this work.

Performance comparison, as in the simulation study of Section \ref{sec_sim_comparsion}, is based on the F1 score at five different budget levels, equally spaced between 0.01 and 0.2.
The results in Figure \ref{fig_real} indicate that the proposed method also outperforms all competitors in this case study and achieves an F1 score greater than 0.90 for a budget greater than 0.152.
In comparison, the \textit{MEWMA} and \textit{unsupervised} methods provide unsatisfactory performance, whereas the \textit{random} and \textit{equispaced} active learning strategies show limited improvements as the budget increases.
This case study underscores the practical value of the proposed method in industrial settings where costs can be reduced through an efficient budget allocation strategy to test the quality of the process.

\section{Conclusions}
\label{sec_conclusions}

This article introduces a novel stream-based active learning method integrated with a partially hidden Markov model (pHMM) to address the limited labeling resource constraint issue in statistical process monitoring applications to monitor data streams characterized by temporal correlation. 
The proposed method is the first to address the modern challenges of online process monitoring, characterized by class imbalances typical of infrequent out-of-control (OC) states and labeling budget constraints.
The integration of the pHMM allows for capturing the temporal dynamics inherent in sequential data. 
At the same time, the stream-based active learning strategy efficiently utilizes labeling resources by focusing on the most informative data points and allows the reveal of new classes corresponding to unseen OC process states. 
Then, the proposed method allows for multiclass classification with an unknown number of classes.

An extensive simulation study demonstrates that the proposed method performs superiorly over alternative approaches, particularly in mid-range budget and shift sizes. 
Competing methods face difficulties in using information from a limited number of labeled samples or distinguishing between the in-control and OC distributions.
The proposed method is shown to be a valuable tool for industrial applications where labeling data points is infeasible on a large scale, such as the ultrasonic tests presented in the case study on the resistance spot welding process in the automotive industry.
The case study shows that costs can be reduced through an efficient budget allocation strategy to test the quality of the process.

Future research can address fine-tuning the exploration and exploitation weight parameters to adapt to specific industrial environments or different process anomalies and investigate the combination with other machine learning classifiers.

\section*{Acknowledgements}
The research activity of A. Lepore was carried out within the Extended Partnership MICS (Made in Italy – Circular and Sustainable) and received funding from the Next GenerationEU of the European Union (PIANO NAZIONALE DI RIPRESA E RESILIENZA (PNRR) – MISSIONE 4 COMPONENTE 2, INVESTIMENTO 1.3 – D.D. 1551.11-10-2022, PE00000004). 
This work reflects only the authors’ views and opinions, neither the European Union nor the European Commission can be considered responsible for them.

\section*{Supplementary Materials}
The Supplementary Materials contain the \textsf{R} code to reproduce graphics and results over competing methods in the simulation study.

\bibliographystyle{apalike}
\setlength{\bibsep}{5pt plus 0.3ex}
{\small
\bibliography{ref}
}

\appendix{}

\section{Simulate From a Partially Hidden Markov Model}

\label{sec_sim_fitted}

Assume that we want to simulate, from a pHMM with estimated parameters $\hat{\bm \theta} = (\hat{\bm \pi}, \hat{\bm{A}}, \hat{\bm b})$, a partially observed sequence of states $x_1, \dots, x_T$, with corresponding emitted observations $\bm y_1, \dots, \bm y_T$.
We assume that we can fix $x_1$ and/or $x_T$, so that we want a sequence of states that possibly starts from a desired initial state $x_1 = i$ and possibly ends in a desired state $x_T = j$.
If $x_1$ is unknown, we first simulate $x_1$ from its probability distribution, which may be conditional on the knowledge of $x_T$.
If $x_1$ is known, we need to simulate $x_2$ given the knowledge of $x_1$ and possibly $x_T$.
After simulating $x_2$, we simulate $x_3$ given the knowledge of $x_2$ and possibly $x_T$, and so on until the end of the sequence.
We need to be able to calculate the following probabilities:
\begin{enumerate}
    \item $p(x_{1} = h)$ in the case $x_1$ and $x_T$ are unknown
    \item $p(x_{t+1} = h | x_t = i)$ in the case $x_t$ is known and $x_T$ is unknown
    \item $p(x_{t+1} = h | x_t = i, x_T = j)$ in the case $x_t$ and $x_T$ are known
    \item $p(x_{1} = h | x_T = j)$ in the case $x_1$ is unknown and $x_T$ is known
\end{enumerate}
Let us consider case by case.

If $x_1$ and $x_T$ are unknown, $p(x_{1} = h) = \pi_h$

If $x_t$ is known and $x_T$ is unknown, $p(x_{t+1} = h | x_t = i) = a_{ih}$

If $x_t$ and $x_T$ are known, the probability to be in state $h$ at time $t$ is:
$$
\tilde{\gamma_t}(h) = p(x_t = h | x_t = i, x_{T} = j) = 
$$
$$
=\frac{p(x_t = i, x_t = h, x_{T} = j)}{p(x_t = i, x_{T} = j)} = 
$$
$$
=\frac{p(x_t = i, x_t = h, x_{T} = j)}{\sum_{k=1}^N p(x_t = i, x_t = k, x_{T} = j)} =
$$
$$
= \frac{p(x_{T} = j | x_t = h) p(x_t = h| x_t = i) p(x_t = i)}{\sum_{k=1}^N p(x_{T} = j | x_t = k) p(x_t = k | x_t = i) p(x_t = i)} =
$$
$$
= \frac{p(x_{T} = j | x_t = h) p(x_t = h| x_t = i)}{\sum_{k=1}^N p(x_{T} = j | x_t = k) p(x_t = k | x_t = i)}$$
We can calculate the probabilities $p(x_{T} = j | x_t = h)$ and $p(x_t = h| x_t = i)$ with a forward-backward algorithm.
\\
\textit{Forward algorithm}: we define $\tilde{\alpha}_t (h) = P(x_t = h | x_t = i)$ for $t = t_1, t_1 + 1, \dots, T$:
\begin{enumerate}
    \item $\tilde \alpha_{t_1}(h) = p(x_t = h | x_t = i) = I(h = i)$, where $I$ is the indicator function,
    \item for $t = t_1 + 1, \dots, T$, 
    $$\tilde \alpha_t(h) = p(x_t = h | x_t = i) = \sum_{k = 1}^N p(x_t = h | x_{t - 1} = k) p( x_{t - 1} = k | x_t = i) = \sum_{k = 1}^N a_{kh} \tilde{\alpha}_{t-1} (k).$$
\end{enumerate}
\textit{Backward algorithm}: we define $\tilde{\beta}_t (h) = P(x_{T} = j | x_t = h)$ for $t = T, T - 1, \dots, 1$:
\begin{enumerate}
    \item $\tilde \beta_{T}(h) = p(x_{T} = j | x_{T} = h) = I(h = j)$,
    \item for $t = T - 1, \dots, 1$, 
    $$\tilde \beta_t(h) = p(x_{T} = j | x_t = h) = \sum_{k = 1}^N p(x_{T} = j | x_{t+1} = k) p( x_{t + 1} = k | x_t = h) = \sum_{k = 1}^N a_{hk} \tilde{\beta}_{t+1} (k).$$
\end{enumerate}
Finally, we can calculate the desired probability: 
$$\tilde{\gamma}_t(h) = p(x_t = h | x_t = i, x_{T} = j) =$$
$$
=\frac{p(x_{T} = j | x_t = h) p(x_t = h| x_t = i)}{\sum_{k=1}^N p(x_{T} = j | x_t = k) p(x_t = k | x_t = i)}=
$$
$$
=\frac{\tilde{\alpha}_t(h) \tilde \beta_t(h)}{\sum_{k=1}^N \tilde{\alpha}_t(k) \tilde \beta_t(k)}.
$$

If $x_1$ is unknown and $x_T$ is known, we only need to modify the first step of the forward algorithm above, i.e., $\tilde \alpha_{t_1}(h) = \pi_h$.

After we have simulated a sequence $x_1, \dots, x_T$, we can simulate $\bm{y}_t$ from the corresponding emission distribution, i.e., if we simulate $x_t = i$, then we simulate $\bm{y}_t$ from $N(\bm \mu_i, \bm \Sigma_i)$.

\section{Initialization of the Partially Hidden Markov Model}
\label{sec_init}

As mentioned in Section \ref{sec_hmm}, the likelihood function is characterized by having multiple local maxima.
Then, the Baum-Welch algorithm, used to iteratively maximize the log-likelihood function starting from a parameter, is prone to converge to a local maximum.
A naive approach to try to reach the global maximum is to use multiple starting random values.
The initialization problem in mixture models such as HMMs has been faced in several works
\citep{shireman2017examining,maruotti2021initialization}.
However, we found that none of the solutions proposed in the literature works well in our case, where the initialization problem is exacerbated by the presence of highly imbalanced classes.
Therefore, the proposed algorithm for initializing pHMMs is able to take advantage of the prior information available on the problem.

To fit a pHMM with $N \geq 2$ states, first, since we assume that the majority of the samples come from an IC process, compute a robust multivariate location and scatter estimate $\hat{\bm \mu}_1$ and $\hat{\bm \Sigma}_1$ of $\bm \mu_1$ and $\bm \Sigma_1$, respectively, such as the Rocke estimator \citep{rocke1996identification}.
These represent the estimated parameters of the pHMM with one state, $N=1$.
For subsequent values of the number of states $N > 1$, we initialize $\bm \pi$ with a vector that gives probability 1 to the first state, i.e., we assume that the stream starts IC, while for the transition probability, we initialize $\bm A$ with diagonal elements equal to 0.99 and off-diagonal elements equal to $0.01/(N-1)$.
To fit the pHMM with $N = 2$, we calculate the moving average $\tilde{\bm Y}_t$ of the observations $\bm Y_t$ with window $k$, and calculate the Mahalanobis distance of these values from $\hat{\bm \mu}_1$.
We take observations $\tilde{\bm y}_{(1)}, \dots, \tilde{\bm y}_{(n_{\text{try}})}$ corresponding to the largest $n_{\text{try}}$ values and, for each $t$ in $1,\dots,n_{\text{try}}$, we fit a pHMM with two states and initialization parameters $\bm \mu_1 = \hat{\bm{\mu}}_1$, $\bm{\mu}_2 = \tilde{\bm y}_{(t)}$, $\bm \Sigma = \hat{\bm \Sigma}_1$.
Finally, we select the model that achieves the maximum likelihood.
For larger values of $N$, we repeat the same procedure as before, where, to obtain the $n_{\text{try}}$ starting values for the mean, for each $t$, we take the smallest Mahalanobis distance from $\tilde{\bm y}_t$ and the estimated means in the previous steps.

To assess the validity of the initialization method, we perform a numerical analysis, under the same conditions of in Section \ref{sec_sim_comparsion}, but limited to $p=20$.
We compare the \textit{proposed} method, with the initialization implemented as described above, against two alternatives.
In the first case, the mean and covariance of each state are initialized to the true values used for data generation.
We denote this method by \textit{true}.
On the one hand, one may think that it should provide the largest likelihood, since it is expected to converge often close to the true parameter values. 
However, this does not guarantee that the solution obtained is a global maximum of the likelihood function.
Therefore, in the comparison, we include another method that uses both our initialization and \textit{true} and then selects the solution that provides the largest likelihood value, which we denote as \textit{both}. 
\begin{figure}[t]
    \centering
    \includegraphics[width=\textwidth]{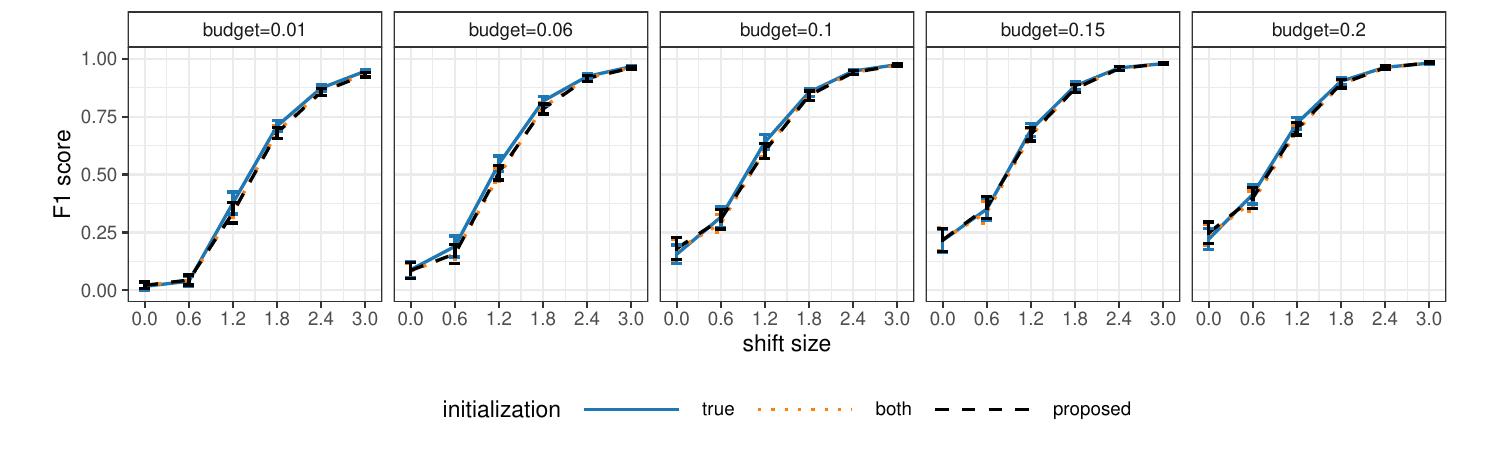}
    \caption{\label{fig_init}
    F1 score plotted as a function of the shift size, for $p=20$ and each available budget. Each line corresponds to a different initialization method.}   
\end{figure}
The results in Figure \ref{fig_init} show that there are no significant differences among all initialization methods in terms of the F1 score.
Only for budget values not greater than 0.1, initializing with \textit{true} parameter values achieves slightly larger F1 scores with respect to the case where \textit{both} our method and \textit{true} parameter values are used.
This means that the estimate obtained with \textit{true} is too optimistic and is not a global maximizer of the likelihood.
On the other hand, the F1 score achieved with the \textit{proposed} method is comparable to the one obtained using \textit{both} initialization methods.
This shows that our method can converge to more realistic estimates with respect to initializing from the \textit{true} values, validating the choice of the \textit{proposed} initialization method.

\section{Additional Simulation Results}

\begin{figure}[t]
    \centering
    \includegraphics[width=\textwidth]{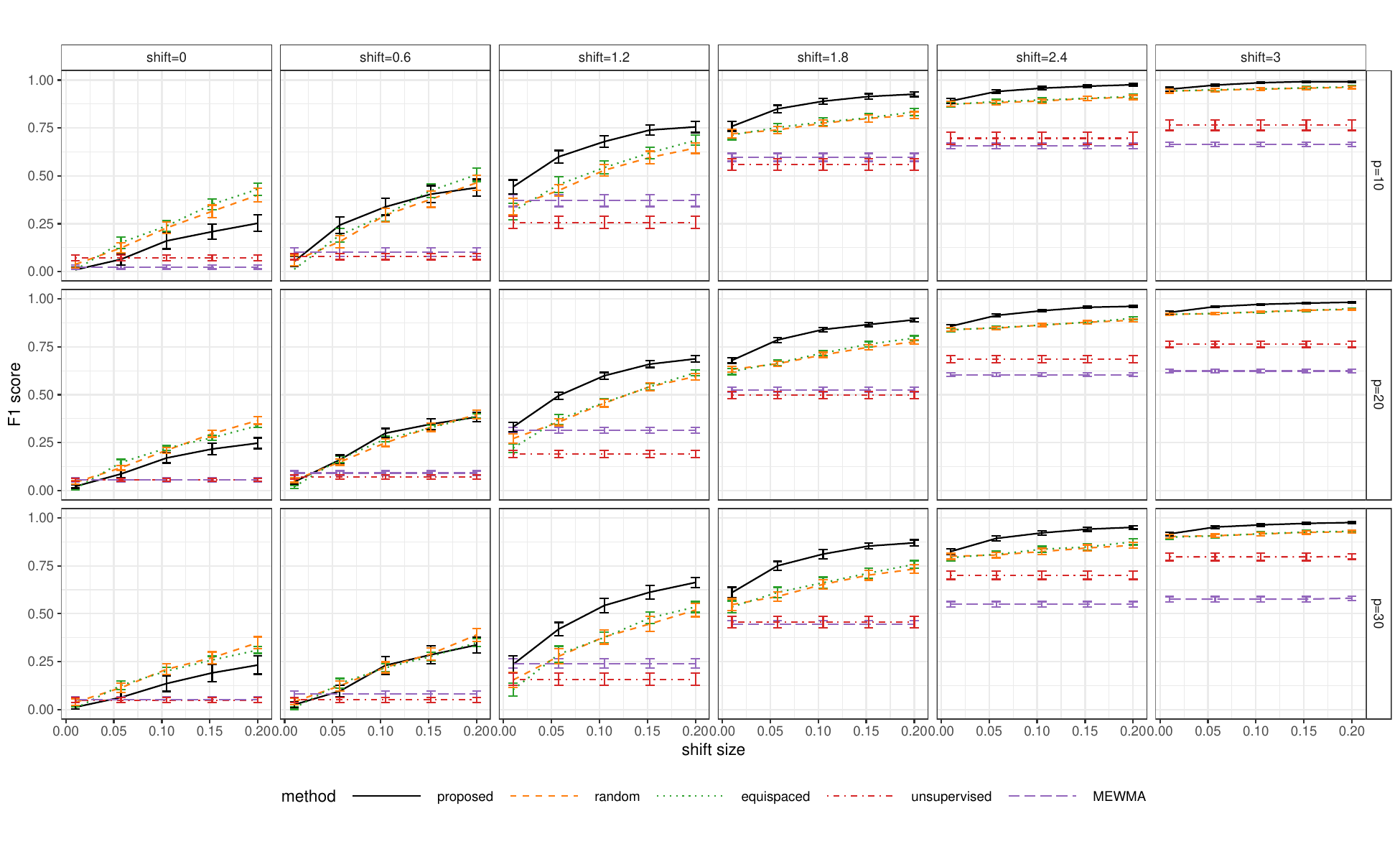}
    \caption{\label{fig_budget}
    F1 score plotted as a function of the available budget for each severity level.}   
\end{figure}

Figure \ref{fig_budget} provides an additional graphical display of the simulation results presented in Section \ref{sec_sim_comparsion}. 
The results are the same as shown in Figure \ref{fig_shift}, where, however, the role of the shift size and the budget is inverted.
That is, each column of panels refers to a shift size, while the F1 score is plotted as a function of the budget.

The main results discussed in Section \ref{sec_sim_comparsion} are further confirmed in Figure \ref{fig_budget}, which allows us to appreciate better the effect of the budget on the performance of the proposed methods. 
In particular, the F1 score for the \textit{MEWMA} and the \textit{unsupervised} methods are constant with respect to the budget, while all active learning methods show an increasing trend.

\end{document}